\documentclass{sigchi}

\makeatletter
\DeclareRobustCommand{\PHP}{%
  \begingroup
  \leavevmode\,\vphantom{P}%
  \dimen\z@=.5\fontcharht\font`P\relax
  \dimen\tw@=0.33333\dimen\z@
  \ooalign{%
    \raisebox{\dimexpr\dimen\z@+2\dimen\tw@-0.4pt}{\rule{\fontcharwd\font`P}{0.4pt}}\cr
    \raisebox{\dimexpr\dimen\z@+\dimen\tw@-0.2pt}{\rule{\fontcharwd\font`P}{0.4pt}}\cr
    P\cr
  }%
  \endgroup
}
\makeatother
\usepackage{multirow, makecell}

\usepackage{balance}       
\usepackage{graphics}      
\usepackage[T1]{fontenc}   
\usepackage{txfonts}
\usepackage{mathptmx}
\usepackage[pdflang={en-US},pdftex]{hyperref}
\usepackage{color}
\usepackage{booktabs}
\usepackage{textcomp}

\usepackage{microtype}        
\usepackage{ccicons}          

\usepackage{todonotes}

\def\plaintitle{Pavement Distress Detection and Segmentation using YOLOv4 and DeepLabv3 on Pavements in the Philippines}

\def\emptyauthor{}
\def\plainkeywords{Pavement Distress; Object Detection; Semantic Segmentation}

\makeatletter
\def\url@leostyle{%
  \@ifundefined{selectfont}{
    \def\UrlFont{\sf}
  }{
    \def\UrlFont{\small\bf\ttfamily}
  }}
\makeatother
\def\pprw{8.5in}
\def\pprh{11in}

\definecolor{linkColor}{RGB}{6,125,233}
\hypersetup{%
  pdftitle={\plaintitle},
  pdfauthor={\emptyauthor},
  pdfkeywords={\plainkeywords},
  pdfdisplaydoctitle=true, 
  bookmarksnumbered,
  pdfstartview={FitH},
  colorlinks,
  citecolor=black,
  filecolor=black,
  linkcolor=black,
  urlcolor=linkColor,
  breaklinks=true,
  hypertexnames=false
}

\makeatletter
\def\@copyrightspace{\relax}
\makeatother
\begin{document}

\title{\plaintitle}

\numberofauthors{1}
\author{%
  \alignauthor{James-Andrew Sarmiento\\
    \affaddr{University of the Philippines Diliman}\\
    \email{jrsarmiento1@up.edu.ph}}\\
}

\maketitle

\begin{abstract}
  Road transport infrastructure is critical for safe, fast, economical, and reliable mobility within the whole country that is conducive to a productive society. However, roads tend to deteriorate over time due to natural causes in the environment and repeated traffic loads. Pavement Distress (PD) detection is essential in monitoring the current conditions of the public roads to enable targeted rehabilitation and preventive maintenance. Nonetheless, distress detection surveys are still done via manual inspection for developing countries such as the Philippines. This study proposed the use of deep learning for two ways of recording pavement distresses from 2D RGB images - detection and segmentation. YOLOv4 is used for pavement distress detection while DeepLabv3 is employed for pavement distress segmentation on a small dataset of pavement images in the Philippines. This study aims to provide a basis to potentially spark solutions in building a cheap, scalable, and automated end-to-end solution for PD detection in the country.
\end{abstract}



\section{Introduction}
\subsection{The Role of Roads in Economic Development}
Road transport infrastructure strengthens the country's economic development indirectly through a multitude of avenues: empowerment of private investment, the formation of new supply chains, or restructuring of economic geography. It is a key sector in the economy of a country in reaching the societal goal of sustained economic growth towards poverty reduction. Among many things, the establishment of road networks provides access to urban centers and rural lands, allows access between key production areas and markets, facilitates the flow of people and products across the country, and enables cost-efficient movement of goods and services. An efficient road transport system is critical for a safe, fast, economical, and reliable mobility within the whole country that is conducive to a productive society.

\subsection{Road Transport Infrastructure in the Philippines}
In the Philippines, road transport accounts for 98\% of passenger transport and 58\% of cargo traffic. The Department of Public Works and Highways (DPWH) is the primary governing body that continuously develops technology and infrastructure facilities for all public works and highways in the country. As of 2011, the country's road system stretches to a total of about 215,000 kilometers where only 78\% of which are comprised of national roads and 18\% of local roads are paved with either asphalt or concrete \cite{adbtransportreport}. The national road paved rate (paved length of total length) remains below the government's original target of 95\% and had slowly risen only from 71\% in 2001, to 78\% in 2011, and 86\% in 2015 \cite{adbtransportreport, aseanstudy2010, aseanstudy2015}. This is significantly lower than other ASEAN countries such as Indonesia (89.7\%), Brunei (93\%), and Singapore (100\%). According to the World Economic Forum, the Philippines garnered a score of 3.7, with 1 being the lowest and 7 being the highest, in the Quality of road infrastructure department which ranks the country as 88th out of 141 countries \cite{worldeconomicforum}. Moreover, the annual investment on the national road system and its quality are lagging in comparison to neighboring countries. As a result, upgrading, expanding, and rehabilitating the road network is limited.

\subsection{Pavement Distress Management and Rehabilitation}
Since pavements tend to deteriorate over time due to environmental conditions and under repeated traffic loads, efficient monitoring of pavement health shall inform road engineers in facilitating effective actions for timely pavement maintenance and immediate repairs. The severity of distress of pavement shall accumulate over time; hence delaying the appropriate actions may lead to instances in which the pavement is beyond patching and should be replaced entirely. 

According to the analysis from Asian Development Bank, for every additional \$1 a developing country spends for road maintenance, road users save \$3 \cite{adb2003road}. On a World Bank report, each dollar spent on patching shall save between \$1.4 to \$44 in operating costs per annual basis \cite{worldbankreport}. In the Philippines, the current allocated budget for repair of national roads and bridges is \PHP11 billion in 2018 and both the road network and budget allocation are expected to grow at a commensurate rate with the establishment of the administration's "\textit{Build, Build, Build}" project. This will result to additional pavement infrastructures which should be evaluated and maintained. Hence, pavement distress detection plays a vital role to identify immediate and appropriate maintenance and rehabilitation needs, monitor pavement conditions over time, develop network preventive maintenance strategies, plan maintenance budget, and procure pavement materials and design. 

\subsection{Manual and Semi-Manual Pavement Distress Detection}
Current pavement distress detection, measurement, and evaluation are done either manually or semi-automated. Manual surveys require manual inspection in the field where the surveyor draws a distress location map and measures specific features such as length of the distress, orientation, and severity \cite{pavementdistressyolo}. Manuals are developed to provide a uniform basis for collecting pavement distress data which contain standards for a common language among pavement defects, classification of distresses into their respective category, and proper severity ratings for each measured defect \cite{fhwamanual}. A manual survey is simple and convenient but expensive and time-consuming which requires expertise from the surveyor. Moreover, this approach involves unreliable and high labor costs, risk of inconsistencies, inconvenient impact on traffic flow, and dangers imposed on the surveyors working on highways.

Pavement distress detection has come a long way from manual inspection to semi-automated methods. Automated methods can be divided into three types according to equipment used: a) Automatic Road Analyzer (ARAN) profile system \cite{aran} and ARRB Hawkeye system \cite{arrb} which employs depth camera or 3D LiDar that uses ultraviolet or near-infrared light to create point cloud image of the pavement \cite{lidar}, b) Ground Penetrating Radar (GPR) and infrared thermal cameras, and c) fixed camera to collect 2D RGB images of the road \cite{gappaper} where the camera module could be mounted in a vehicle or handheld to capture road images manually. 

The first method can detect noticeable changes in height for defects such as potholes, while the second can be utilized for detection of deep cracks, evaluation of the magnitude of the flaw, layers separation, and fault leveling due to the temperature change occurring between cracks and asphalt surface \cite{gprthermal}. Both types provide more information from the pavement images but are expensive, require more work to mount in a vehicle, and require huge processing power for its massive data outputs. The third type proves to be more economical, flexible, and efficient using 2D RGB images. However, there are challenges with using 2D pavement images, such as varying image source, crack non-uniformity, insufficient background illumination, and presence of unrelated background subjects. 

\subsection{Deep Learning in PD Detection}
Deep Learning has brought remarkable breakthroughs in speech recognition, visual object recognition, image classification, object detection, semantic segmentation, and other domains \cite{deeplearninghinton}. It has emerged as a powerful technique for learning feature representation that enables localization of object instances in the image data \cite{deeplearningpavementreview}. Since it is now efficient to collect pavement distress data using 2D RGB images, using deep learning in pavement distress detection provides an automated solution in detecting defects and cracks on public roads. 

Other countries have already employed semi-automated methods to collect data and evaluate their public roads and pavements, while the Philippines has been lagging while the Philippines has been lagging through its manual inspection as primary method for PD identification and evaluation. In this paper, the author proposed the use of YOLOv4 and DeepLabv3 to identify distresses from images captured via a smartphone on pavements in the Philippines. The capability of using deep learning to detect pavement distresses makes it possible to build an end-to-end solution that would enable scalable and semi-automated road surveys for condition monitoring system for the country. 

\section{Pavement Distress Detection}
Development of vision-based pavement distress detection revolves around popular methods using intensity-thresholding, edge detection, wavelet transform, traditional machine learning techniques, and deep learning. 

Previous studies involving intensity-thresholding \cite{intensity-thresholding-dynamic, intensity-thresholding-neighboring, intensity-thresholding}, edge detection techniques \cite{edge-detection-bemd, edge-detection-probablitymaps, edge-detection-crack-tree, edge-detection-fuzzylogic}, wavelet-transforms \cite{wavelet-trous, wavelet-chiang}, and traditional machine learning methods such as Support Vector Machines (SVM) \cite{mlsvm, mlcomprehensivereview}, and shallow Artificial Neural Networks (ANN) \cite{mlmethodscrack, mlann} are quite effective on small scale distress detection focusing only on pavement cracks. However, these techniques are sensitive to noisy pixels on images, heavily reliant to image orientation, unable to distinguish various types of distresses, and require manual extraction of features. 

Deep Convolutional Neural Networks (DCNN) demonstrated success in large scale image classification and object recognition of the images without manual feature engineering by being able to understand extract complex, high-level feature mapping from raw input data. \cite{BengioDL}. Deep Learning architectures are inspired by how a mammalian brain works in using different areas of the cortex to abstract different levels of features when given an input. When given an input precept, the mammal brain processes it using different areas of the cortex which abstracts different levels of features. \cite{deeplearningvisualprocessing}.

DCNNs are shown to be highly effective in processing images and videos by taking the raw input at the lowest level and processing them through a sequence of computations to obtain feature representations and feature maps for classification and detection in the higher layers \cite{deeplearningfeatures}. This is done by passing the image through a series of convolutional layers, pooling layers, and fully-connected layers to build a multitude of feature maps that describe the image. A deep learning architecture means it has multiple hidden layers in the network. In contrast, a shallow architecture has only a few hidden layers (1 to 2 layers). In deep learning, multiple hidden layers learn many levels of abstraction that a shallow architecture could not learn \cite{deeplearningvisualprocessing}.

Deep learning (DL) has attracted numerous research as a successful alternative approach to image-based distress detection with at least 12 recently published papers from 2016 to 2018 \cite{deeplearningpavementreview} as this approach is producing unprecedented results in both detecting and classifying pavement faults. Table \ref{tab:distress_studies} shows the summary of surveyed papers using deep-learning in this paper. Currently, published papers can be broadly divided into two categories depending on whether the classification or detection happens on a block-level or image-level. 

\begin{table}
  \caption{Pavement Distress Detection Literatures}
  \label{tab:distress_studies}
  \begin{tabular}{l l l l l}
    \toprule
     Reference & Method & Classes & Level \\ 
    \midrule
    Wang \cite{gridbasedsmartphone} & CNN + PCA & 1 &  Block  \\
    Eisenbach \cite{gappaper} & CNN & 6 &  Block  \\
    Zhang \cite{crack-det1-zhang} & CNN & 1 & Image  \\
    Gopalakrishnan \cite{transferlearningpavementdistress} & CNN & 1 & Image \\
    Lau \cite{crack-unet} & U-Net & N/A & Block \\
    Escalona \cite{crack-unet} & U-Net & N/A & Block \\
    Some \cite{automaticroadcracksome} & CrackIt & 1 & Image \\
    Du \cite{pavementdistressyolo} & YOLO & 7 & Image \\
    Maeda \cite{smartphonedistressfront} & SSD & 6 & Image \\
    Lei \cite{baidustreetview} & YOLOv3 & 8 & Image \\
    Pereira \cite{unetpothole} & U-Net & N/A & Image \\
  \bottomrule
\end{tabular}
\end{table}

Block-level classification or detection segments a high-resolution image into smaller patches and applies the prediction on each image grid. Wang and Hu \cite{gridbasedsmartphone} pre-processed the image captured by a smartphone camera transforming it into a gray image of 960x704 pixels and segmenting the input into 15x11. They employed CNN to detect the existence of pavement crack in corresponding scale grids so the image only keeps the grids with cracks similar to a crack network as an output. Eisenbach et. al \cite{gappaper} used a Deep Convolutional Neural Network (DCNN) on their project ASINVOS on large open-source pavement distress recorded by S.T.I.E.R which is a measuring vehicle designed for large-scale pavement condition surveys. ASINVOS takes an input image divided into blocks of three per column and processes each grid to produce an output mask of detected pavement defects including are cracks, potholes, inlaid patches, applied patches, open joints, and bleeding. 

Image-level classification or detection for pavement distress images predicts bounding boxes from the full image resolution localizing the detected crack alongside with the predicted classification. Zhang et. al \cite{crack-det1-zhang} used a low-cost smartphone to capture pavement images and applied a Deep ConvNet trained on square image patches for the classification of patches with and without cracks which outperformed SVM and Boosting methods in terms of F1-score \cite{crack-det1-zhang}. Gopalakrishnan et. al \cite{transferlearningpavementdistress} classified pavement surface images from the FHWA/LTPP as crack or non-crack by using a pre-trained VGG-16 DCNN on ImageNet to generate semantic image vectors fed to SVM with results yielding 0.90 F1 score. Lau et. al \cite{crack-unet-resnet} proposed using U-Net based on pre-trained ResNet-34 architecture while Escalona et. al \cite{crack-unet} implemented three different U-Net models for segmenting cracks in pavement images and scored 96\% and 97.31\% respectively on CrackForest Dataset (CFD) \cite{crack-det-randomforest}.

\begin{figure*}
  \centering
  \includegraphics[width=0.8\linewidth]{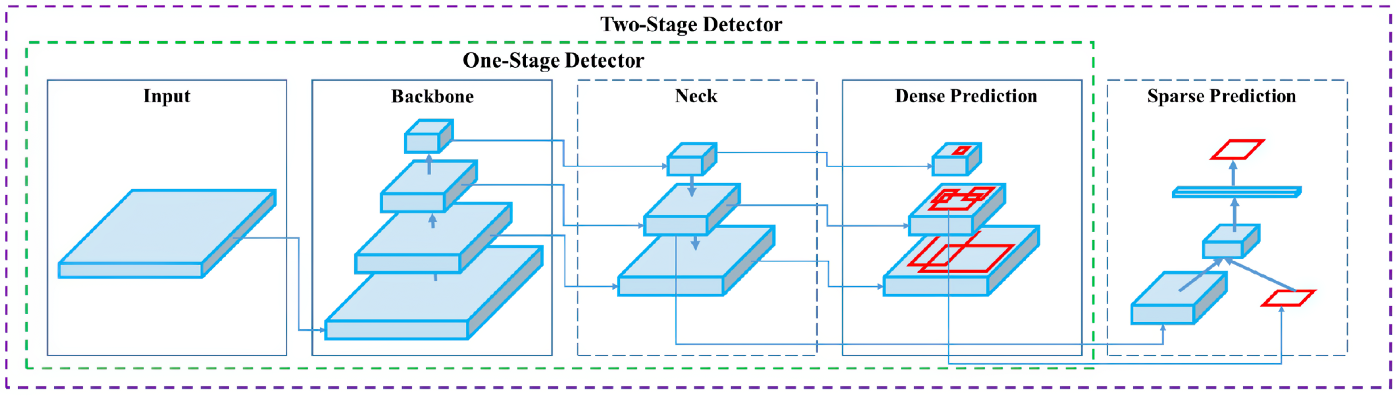}
  \caption{Object Detector Architecture as adopted from YOLOv4}
  \label{fig:objectdetector}
\end{figure*}

Other researches have focused on making image-level classification and detection scalable using street-view images captured by mounting a fixed camera on a vehicle. Some et. al \cite{automaticroadcracksome} mounted the camera on top of the vehicle and then utilized DL to remove features of non-interest (such as cars, sidewalks, surfaces with no distresses, etc.) and detect pavement distresses. Du et. al \cite{pavementdistressyolo} adopted a YOLOv3 network to detect and predict the location of possible distresses and their respective categories on images collected by high-resolution industrial cameras installed on the back of the vehicles. Their model \cite{pavementdistressyolo} has an accuracy of 73.64\% but still faces difficulties when the image has strong or insufficient light or the main information of the pavement is covered by a shadow. Maeda et. al \cite{smartphonedistressfront} created an end-to-end application based on single-shot multibox detector (SSD) with Inception V2 and MobileNet V2 as the backbone, for detecting and classifying damages in road images acquired using a smartphone installed in a moving camera that has an inference time of 1.5s and can achieve recalls and precision greater than 75\% on 300x300 resolution. Pereira et. al proposed a different approach in detecting pavement distresses by using semantic segmentation to segment paved road and pothole with U-Net over smartphone images \cite{unetpothole}.

\section{DPWH Dataset}

\subsection{Data Collection}
\begin{figure}[h!]
  \centering
  \includegraphics[width=0.5\textwidth]{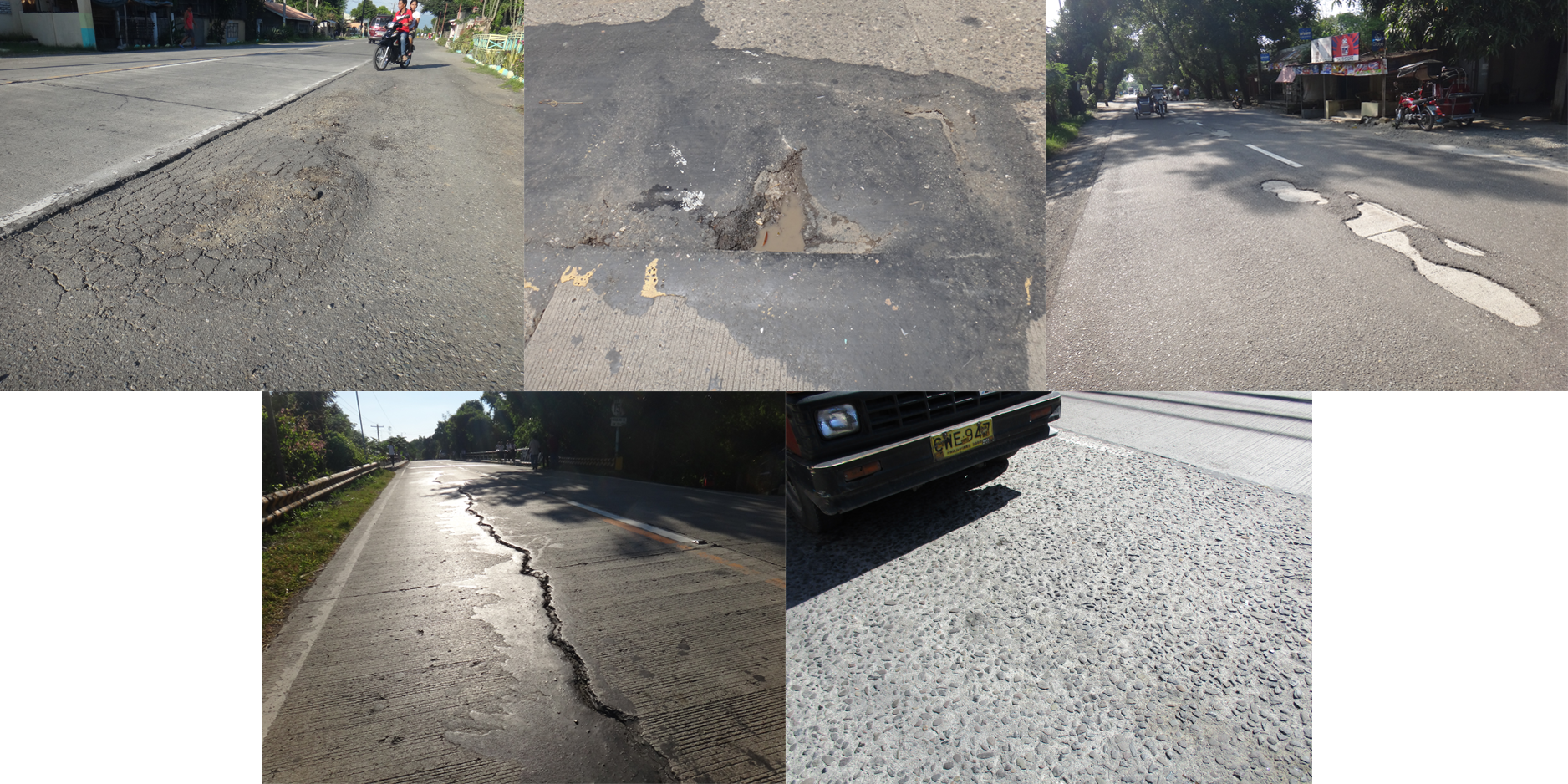}
  \caption{DPWH dataset: Alligator, Bowl-shaped depression, Delamination, Crack, Scaling (from left to right)}
  \label{fig:dpwhdataset}
\end{figure}

The primary dataset is collected by a road engineer surveyor from DPWH using a handheld smartphone. In this data collection task, 305 images were collected with resolution reduced to 900 x 600 to speed up training. For both pavement distress detection and segmentation, the data is divided into 80-20 train-validation split for 242 training set and 63 for validation set. Images are captured with no standardized orientation; thus there is a large variety of how the photos are taken as seen in Figure \ref{fig:dpwhdataset}. 

The data is composed of five most common defects that can be observed on pavements in the country including bowl-shaped depression, delamination, cracks, alligator cracks, and scaling. Bowl-shaped depressions and delaminations are kinds of potholes where the former is a result of the loss of wearing course or base materials while the latter is the loss of a discreet area of a wearing course layer caused by poor bonding between the surface layer and the lower layer. Cracks are breaks in the pavement surface formed by physical tension which occur in a variety of patterns, ranging from isolated single crack to an interconnected pattern that could extend over the entire pavement surface. Fatigue cracking or alligator cracking is the next stage for wheel path longitudinal cracking that creates an interconnection of longitudinal, transversal, and diagonal cracks forming a series of small polygons that resembles an alligator skin \cite{automaticroadcracksome}. Lastly, scaling is the deterioration of the upper concrete slab of more than 10m. Figure \ref{fig:dpwhdataset} shows a sample of each class from the dataset. Each image may have more than one pavement distresses and the distribution of annotations for 305 images are shown in Table \ref{tab:annotations}

\begin{table}
  \caption{Dataset from DPWH}
  \label{tab:annotations}
  \begin{tabular}{l l l l l}
    \toprule
    \multirow{2}*{Class} & \multicolumn{2}{c}{Training} & \multicolumn{2}{c}{Validation} \\
    & Image & Annotation & Image & Annotation \\
    \midrule
    Alligator Crack & 35 & 89 & 9 & 16 \\
    Bowl Depression & 48 & 110 & 13 & 22  \\
    Delamination & 40 & 525 & 10 & 147 \\ 
    Crack & 59 & 214 & 15 & 58 \\ 
    Scaling & 60 & 253 & 16 & 72 \\
  \bottomrule
   \textbf{Total} & 242 & 1191 & 63 & 315
\end{tabular}
\end{table}

\subsection{Data Annotation}
Both object detection and semantic segmentation tasks require a corresponding ground truth to train a model. A road engineer annotated the images using HyperLabel and PixelAnnotationTool. HyperLabel is the software used to annotate the images with bounding boxes for pavement distress detection task. PixelAnnotationTool is a pixel-wise annotation tool used to generate the ground truth masks for semantic segmentation \cite{pixelannotationtool}. Annotating pavement defects with bounding boxes proved to be a difficult task as the five distresses vary greatly in their appearance. Alligator cracks, delaminations, and bowl-shaped depressions require simple annotation and are usually uniform in sizes. However, cracks and scaling are complicated to annotate as cracks tend to cover a small proportion of the bounding boxes while scaling, when present, usually have a coverage of more than 50\% of the image.

\section{Pavement distress detection with YOLOv4 network}
You Only Look Once (YOLO) is a deep CNN for the purpose of detecting objects from images \cite{yolov1}. It divides the image into a grid of multiple cells and predicts the probability of having an object inside the anchor boxes. This method processes object detection as a regression problem to spatially separate bounding boxes and associated class probabilities from full images in one evaluation using a single neural network. YOLO has a single network for the whole detection pipeline so it can be optimized end-to-end directly on detection performance that enables fast inference.

YOLO framework is a family of one-stage object detectors that had evolved from its first version \cite{yolov1}, to YOLO9000 or YOLOv2 \cite{yolov2}, YOLOv3 \cite{yolov3}, until the most recent YOLOv4 \cite{yolov4}. YOLOv3 was an incremental upgrade to YOLO9000 and YOLOv4 in comparison is a significant upgrade to its predecessors in terms of accuracy (average precision) and speed (FPS), the two metrics generally used to evaluate an object detection algorithm. YOLOv4 was designed to have a fast operating speed object detector for production systems which is also optimized for parallel computations.

\subsection{Overall Architecture of YOLOv4}
The overall network architecture of an object detector is typically composed of several components is shown in Figure \ref{fig:objectdetector}. The family of YOLO frameworks is one-stage detectors while there are two-stage detectors like R-CNN, fast R-CNN and faster R-CNN which are accurate but slow.

\textbf{Input}. This is where to input the image data.

\begin{figure}[h!]
  \centering
  \includegraphics[width=0.3\textwidth]{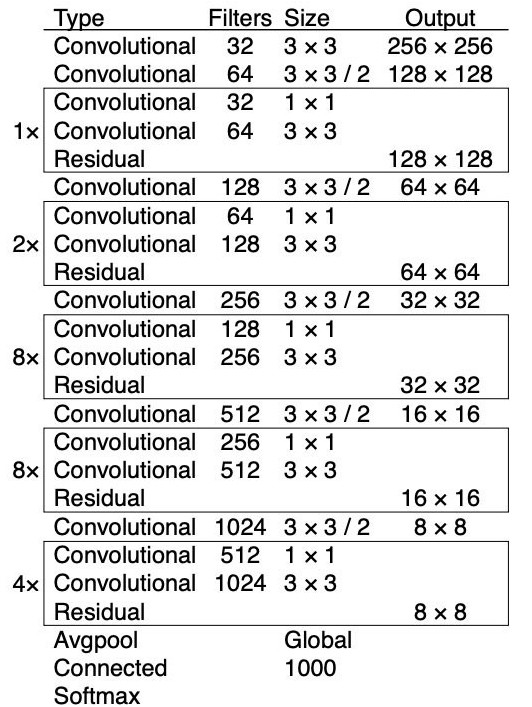}
  \caption{Darknet-53}
  \label{fig:darknet}
\end{figure}

\textbf{Backbone}  Models such as VGG16, DenseNet, ResNet, and DarkNet, etc. are feature-extraction architectures. They are pre-trained on image classification datasets like ImageNet\cite{imagenet} and fine-tuned on a detection dataset. A backbone network takes an input image and produces different levels of feature maps with higher semantics as the network gets deeper. YOLOv4 used CSPDarknet53 as a backbone as it was shown to be the most optimal model. CSP (Cross-Stage-Partial) connections try to enhance the learning capability of CNNs by separating the current layer into 2 parts, one that will go through a block of convolutions, and one that would not \cite{cspnet}. The backbone of YOLOv4 \cite{yolov4} is shown in Figure \ref{fig:darknet}.

\textbf{Neck}. The neck and head are subsets of the backbone with the primary purpose of enhancing feature discriminability and robustness using methods like Feature Pyramid Networks\cite{fpnnetwork}, Path Aggregation Network \cite{pannet}, and Bi-FPN \cite{bifpn}. YOLOv3 uses FPN while YOLOv4 uses a combination of modified Path Aggregation Network (PANet) \cite{pannet}, a modified Spatial Attention Module (SAM) \cite{samnet}, and Spatial Pyramid Pooling (SPP)\cite{spp} as methods for parameter aggregation for adding information in a layer. 

\textbf{Head}. The head block is used to locate bounding boxes and classify bounding boxes. The network detects bounding box coordinates (x, y, width, height) and confidence score for each class. YOLO is an anchor-based object detector which applies the head network to each anchor box. YOLOv4 utilizes YOLOv3 for the head block. 

\textbf{Sparse Prediction} This layer is used in two-stage-detection algorithms such as Faster-R-CNN \cite{fasterrcnn}, etc.

\subsection{YOLOv4 Training Optimizations}
YOLOv4 distinguishes between two additional methods that improve the object detector’s accuracy to achieve a fast operating-speed neural network with good accuracy. 

\begin{figure*}
  \centering
  \includegraphics[width=0.8\linewidth]{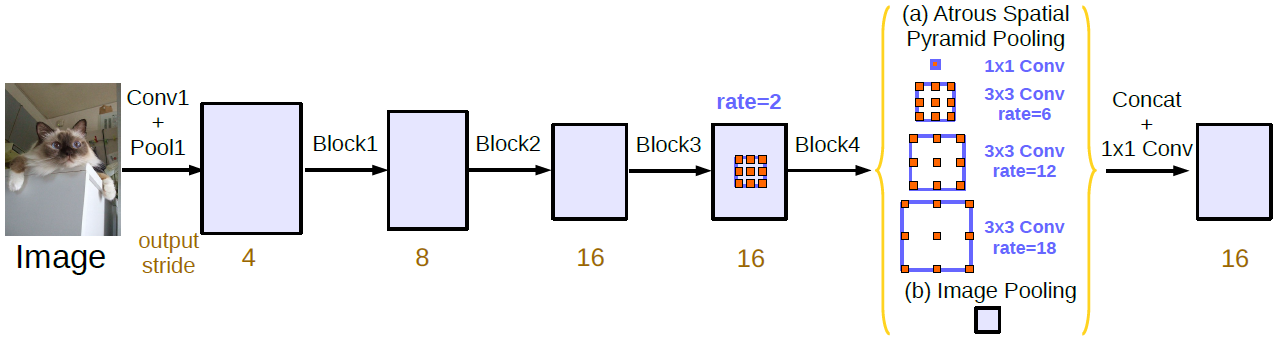}
  \caption{Architecture of DeepLabv3}
  \label{fig:archideep}
\end{figure*}

\textbf{Bag of Freebies}.
Bag of freebies (BoF) are methods that could improve the object detector's accuracy without increasing inference cost by either changing the training strategy or increasing training cost \cite{yolov4}. Data augmentation is a classic example. Augmenting the data generates a new set of data by increasing the variability of original image data so that the model is more robust to the images obtained from various environments.

\textbf{Bag of Specials}
Bag of Specials (BoS) is plugin modules and post-processing methods for enhancing certain attributes in the model. This causes a slight increase in inference cost but significant improvement in the accuracy of the detection model. 

Bag of Freebies and Bag of Specials methods are implemented on the backbone and detector layer. Table \ref{tab:bosbof} illustrates a summary of BoF and BoS enhancements and modules implemented in YOLOv4.

\begin{table}
  \caption{Summary of BoF and BoS used in YOLOv4}
  \label{tab:bosbof}
  \begin{tabular}{c l l}
    \toprule
     & Backbone & Detector  \\ 
    \midrule
    \multirow{9}*{BoF} & CutMix & CIoU-loss \\
    & Mosaic data augmentation & Cross-Mini BatchNorm \\
    & DropBlock & DropBlock \\
    & Class label smoothing & Mosaic data augmentation \\
    & & Self-Adversarial Training \\
    & & Multiple anchors single GT \\
    & & Cosine annealing scheduler \\
    & & Optimal hyperparameters \\
    & & Random training shapes \\
    \hline
    \multirow{5}*{BoS} & Mish activation & Mish Activation \\
    & CSP & SPP-block \\
    & \multirow{2}*{\shortstack[l]{Multi-input weighted \\ residual connections}} & SAM-block \\
    & & PAN path-aggregation block \\
    & & DIoU-NMS \\
  \bottomrule
\end{tabular}
\end{table}

\subsection{YOLOv4 Training Procedure}
\textbf{YOLOv4}. This study used an available YOLOv4 Darknet on Github by AlexyAB \cite{githubdarknet} with a backbone pretrained on MS COCO. The Microsoft COCO dataset involves over 2 million well-labeled objects in 80 various groups with over 300,000 images. The pre-trained weights in the COCO dataset were used to initiate the detection task in the newly proposed models.

\textbf{Hardware} The training process ran on a machine that has a Xeon(R) E5-2620 v3 CPU with 16GB RAM and an RTX 2080Ti GPU with 11GB VRAM.

\textbf{Training Process and Batching}. The model is trained with a batch size of 64 and subdivisions of 32 with 10,000 epochs. The training process ran for 12 hours.

\section{Pavement Distress Semantic Segmentation with DeepLabv3}
Semantic segmentation is the task of clustering parts of images together which belong to the same object class \cite{semanticsegmentation}. The goal is to label each pixel of an image with a corresponding class in which the output is a high-resolution image (typically of the same size as input image). However, one of the challenges in segmenting objects in images using deep convolutional neural networks is that input feature maps become smaller while traversing through the convolutional and pooling layers of the full network which means information about objects of a smaller scale can be lost causing low-resolution predictions and object boundaries being fuzzy. DeepLabv3 addresses this challenge by using Atrous convolutions and Atrous Spatial Pyramid Pooling (ASPP) modules \cite{deeplabv3}.

\subsection{From DeepLabV1 to DeepLabv3}
DeepLabv3 attempts to extend prior research on how to handle objects in images of varying scale to achieve significant improvement over previous DeepLab versions. It has evolved over two generations. DeepLabV1 attempts to use Atrous Convolution and Fully Connected Conditional Random Field (CRF) to capture fine edge details while also catering for long-range dependencies among feature maps \cite{deeplabv1}. DeepLabV2 applies Atrous Spatial Pyramid Pooling (ASPP) to segment objects at different scales by probing an incoming convolutional feature layer with filters at numerous sampling rates and effective fields-of-views capturing objects as well as image context at various scales \cite{deeplabv2}. The latest generation, DeepLabv3, employs atrous convolution in cascade to capture multi-scale context by adopting multiple atrous rates combined with an improved ASPP module with image-level feature encoding to boost performance even without DenseCRF post-processing used on the previous generations \cite{deeplabv3}.

\textbf{Atrous Convolutions}
\newline
Atrous convolution (dilated convolutions) uses an atrous/dilation rate to make field-of-view of filters larger without impacting computation or number of parameters by inserting zeros between two successive filter values along each spatial dimension \cite{dilatedconv}. It allows repurposing very deep convolutional networks to extract denser features maps by removing downsampling operations from the last few layers and upsampling the corresponding filter kernels. 

\textbf{Atrous Spatial Pyramid Pooling}
\newline
ASPP is a network used to obtain multi-scale context information on top of the feature maps extracted from a backbone. Four parallel atrous convolutions with different atrous rates are applied to handle segmenting the object at different scales combined with image-level features to incorporate global context information by applying global average pooling on the last feature map of the backbone \cite{deeplabv2}. The results of each operation along the channel is concatenated, and 1 x 1 convolution is applied to get the output.

\begin{figure*}
  \centering
  \includegraphics[width=1.0\linewidth]{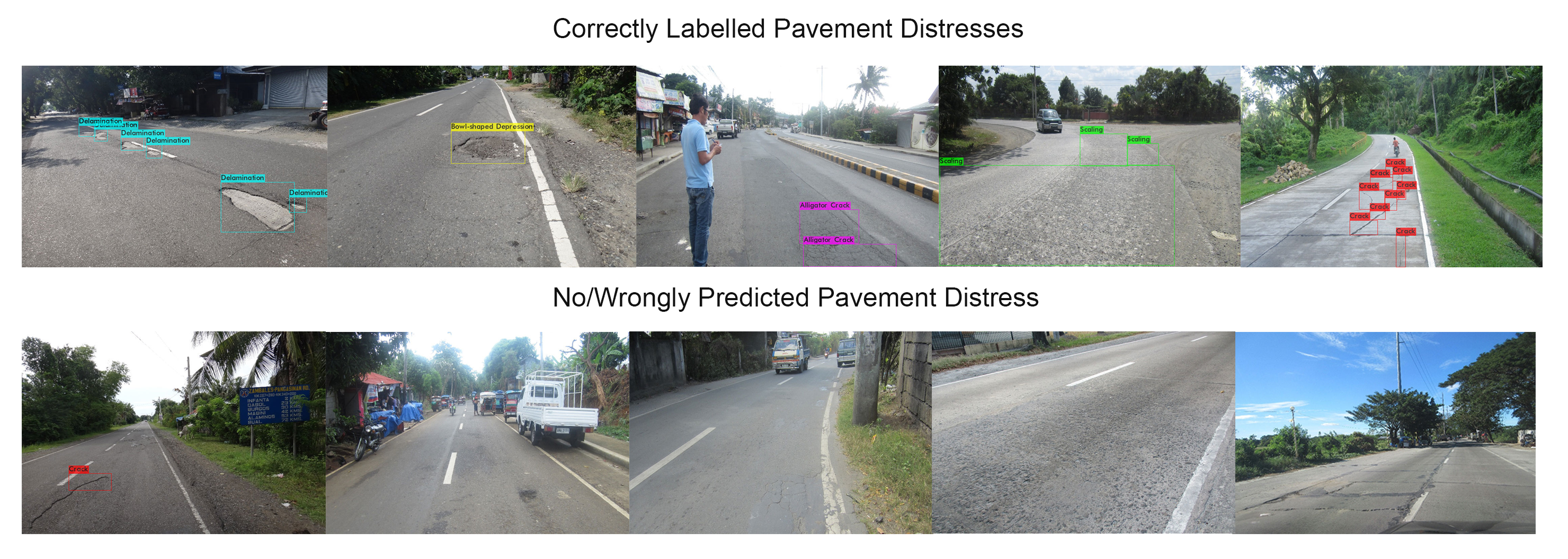}
  \caption{YOLOv4 predictions on sample validation data arranged from highest individual average precision per class}
  \label{fig:yolopredictions}
\end{figure*}

\subsection{Overall Architecture of DeepLabv3}
DeepLabv3 consists of the following architecture: 

\begin{itemize}
    \setlength\itemsep{0.2em}
    \item A backbone network (VGG, DenseNet, ResNet) for feature extraction
    \item An atrous convolution in the last few blocks of the backbone
    \item An ASPP Network to classify each pixel with their corresponding pixel-level classification
    \item A 1 x 1 convolution to get the actual size of the image as an output mask
\end{itemize}

This architecture, as seen in Figure \ref{fig:archideep}, has significant improvements in extracting dense feature maps for long-range contexts and capturing objects at multiple scales by increasing the receptive field exponentially without losing the spatial dimension and improves performance on segmentation tasks \cite{deeplabv3}.

\subsection{Training Procedure}
\textbf{DeepLabv3}. The DeepLabv3 model used for pavement distress segmentation has a pre-trained ResNet-18 backbone on top of the ASPP network.

\textbf{Hardware} The training process ran on a machine that has a Xeon(R) E5-2620 v3 CPU with 16GB RAM and an GTX 1080Ti GPU with 11GB VRAM.

\textbf{Training Process and Batching}. The model is trained with a batch size of 4 for both training and validation sets. The deep learning framework used is PyTorch. The benchmark model is trained for 40 epochs with parameters including weighted cross-entropy as loss function, Adam as an optimizer, and learning rate of 0.0005. In order to reduce overfitting, data augmentations are applied to the dataset which includes ShiftScaleRotate and HorizontalFlip, and RandomBrightnessContrast. The training ran for 8 hours. 

\section{Experimental Results and Discussion}
\subsection{Performance Evaluation for YOLOv4}
\textbf{Pre-processing}. Image-thresholding techniques are effective on pavement cracks because alterations on the pavements cause change in pixel intensities on the image. Histogram equalization is a method of improving contrast globally. Contrast Limited Adaptive Histogram Equalization (CLAHE) is a variant of adaptive histogram equalization to reduce this problem of noise amplification in the image \cite{clahe}. However, this method changes the color of the image into its grayscale equivalent. Two experiments on the DPWH dataset using YOLOv4 uses one without pre-processing and one with CLAHE. 

\begin{figure}[h!]
  \centering
  \includegraphics[width=0.4\textwidth]{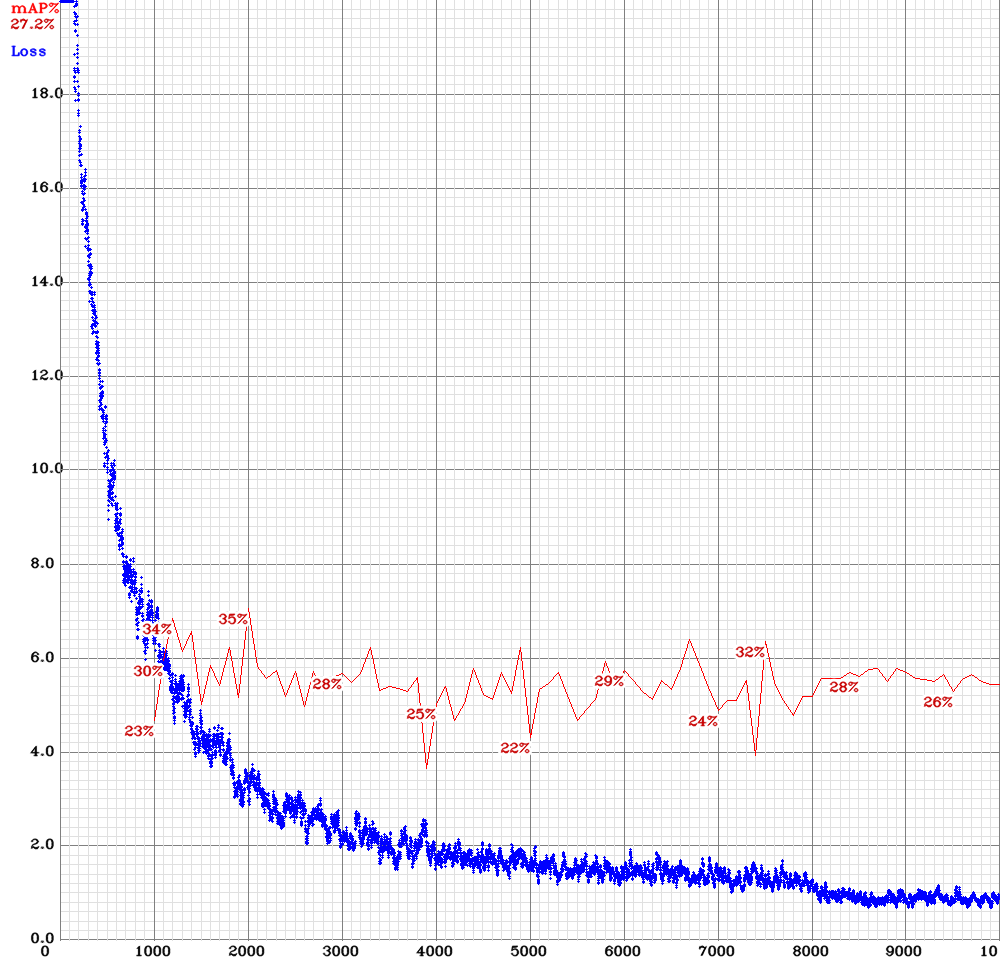}
  \caption{Plot of YOLOv4 Loss and mAP}
  \label{fig:yololoss}
\end{figure}

\subsubsection{Metrics: IoU and mAP}
YOLO generates bounding boxes, predicted classes, and predicted confidences for every image fed into the network. The robustness and accuracy of the network are evaluated by metrics including IoU and mAP.

\textbf{Intersection over Union}. IoU measures the overlap between the ground truth and the predicted bounding box given by the ratio of the area of intersection and area of union. A higher IoU indicates accurate bounding boxes drawn by the model on the detected object. In some datasets, an IoU threshold (e.g. 0.5) is predefined to classify whether the prediction is a true positive or a false.
\begin{center}
   $IoU = \frac{Area of Overlap}{Area of Union}$ 
\end{center}

\textbf{Mean Average Precision}.AP (Average precision) computes the average precision value for recall value over 0 to 1, commonly known as area under the curve. mAP is the mean of AP scores for all classes. In addition to mAP, a simple average F1 score with the AP value per class included to assess the performance of the model on the whole dataset. The formulae for precision, recall, and F1 score are as follows:

$precision=\frac{TP}{TP+FN};recall=\frac{TP}{TP+FN};F=2.\: \frac{precision\: .\: recall}{precision+recall}$

\subsubsection{YOLOv4 on DPWH dataset}
The training procedure took 12 hours on the machine with 10,000 iterations. The mAP on the validation stagnated around mid-20\% (0.25 mAP) and did not go any further than that as seen in Fig. \ref{fig:yololoss}. This is a sign of overfitting in the network that might be caused by the small dataset size. While YOLOv4 uses data augmentation, a sufficient training data should be needed to acquire decent results.

The summary of the performance evaluation of using YOLOv4 on DPWH dataset can be seen in Table \ref{tab:dpwhyolov4}. The network did not score very well on all metrics - mAP, F1 score, and IoU. However, a closer look at the individual AP scores illustrates a wide discrepancy among five classes. 

\begin{figure}[h!]
  \centering
  \includegraphics[width=0.5\textwidth]{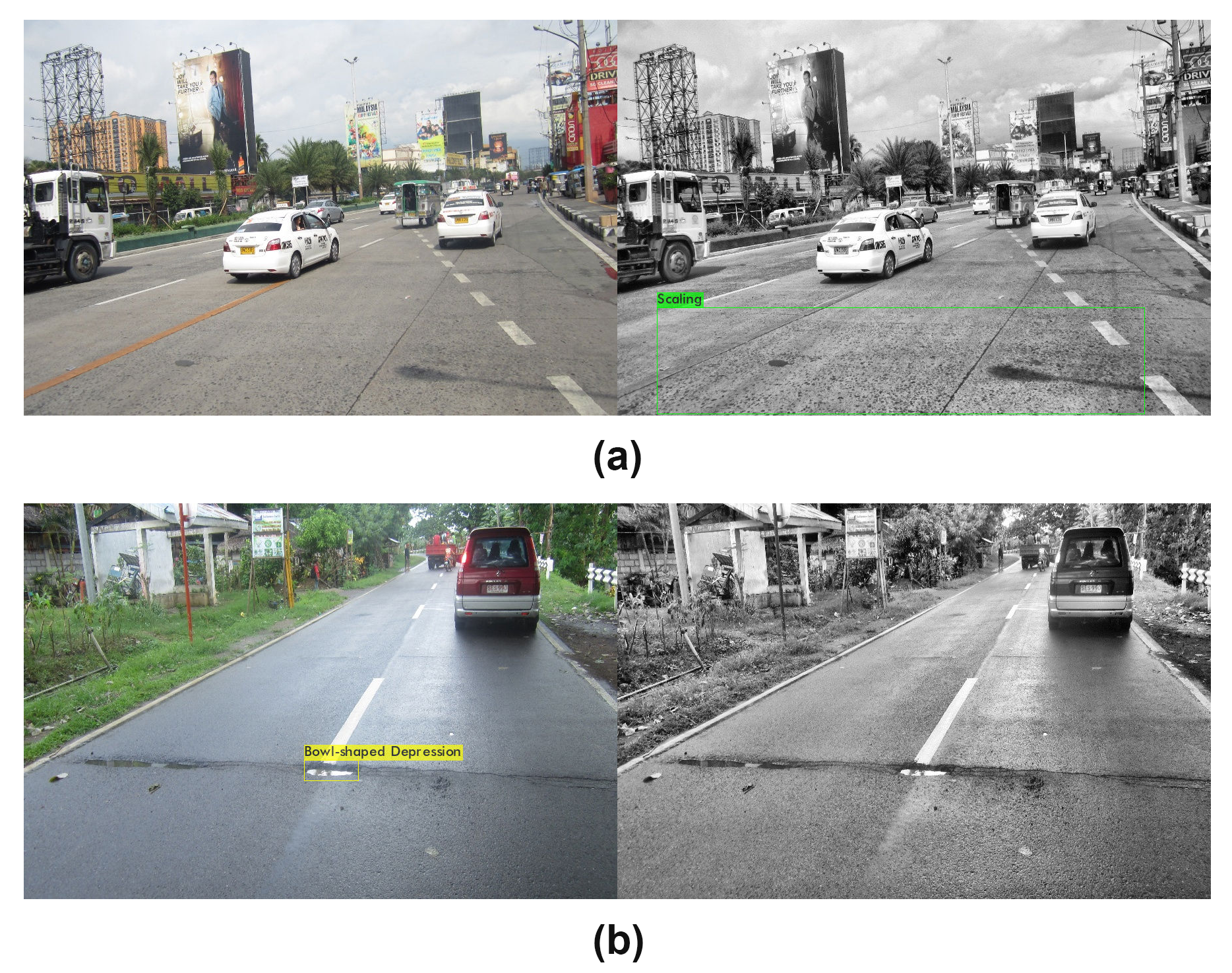}
  \caption{(a) distress detected only on processed image, (b) distress detected only on original image}
  \label{fig:clahe}
\end{figure}

\begin{figure*}
  \centering
  \includegraphics[width=1.0\linewidth]{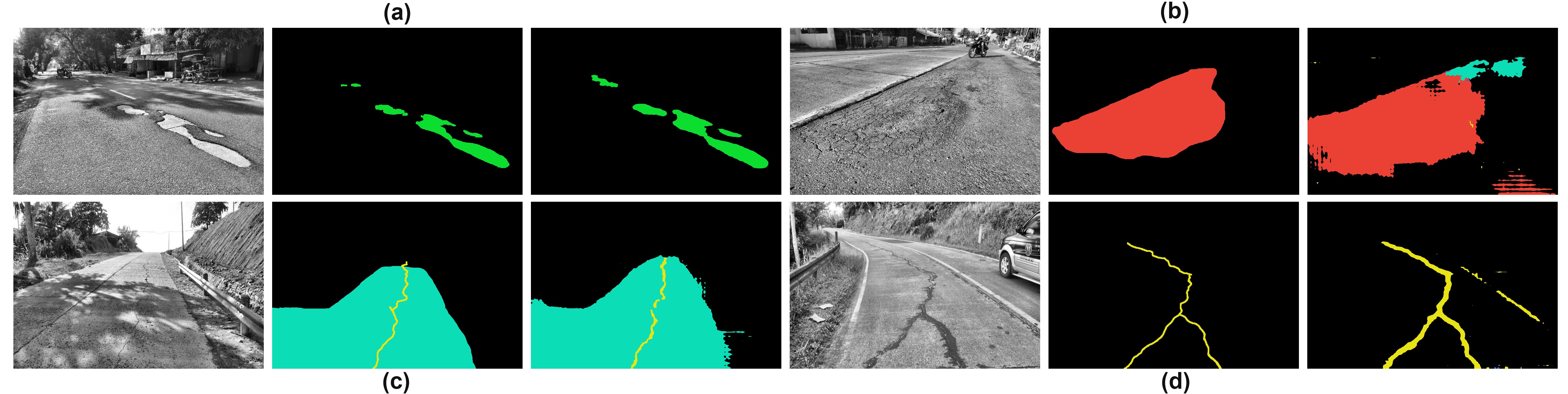}
  \caption{Distress Segmentation (Image, Mask, Prediction). (a) Delamination (green), (b) Alligator mask (red), (c) Scaling (light blue), (c, d) Crack (yellow)}
  \label{fig:deeplab}
\end{figure*}

Figure \ref{fig:yolopredictions} presents YOLOv4 predictions on some data from the validation set. Factors that seemingly play a role in the absence and presence of predictions are the orientation of the camera, the contrast of the image, and the size of the road defects. The orientation of the camera affects viewing angles of the faults leading to variations of the appearance of distresses. The contrast of the image influences the visibility of cracks and scaling in the pavement. The size of road defects influences the network's capability to detect the fault. 

\begin{table}
\centering
  \caption{F1-score and mAP for YOLOv4 on DPWH dataset}
  \label{tab:dpwhyolov4}
  \begin{tabular}{l l l}
\toprule
\textbf{Distress}  & \textbf{Original} & \textbf{Processed} \\
\midrule
    Alligator Crack & 0.306 & 0.369   \\
    Bowl Depression & 0.327 & 0.213   \\
    Delamination & 0.624 & 0.711   \\
    Crack & 0.038 & 0.05    \\
    Scaling & 0.065 & 0.087    \\
\midrule
\textbf{mAP}  & 0.272    & 0.286    \\
\textbf{F1-score} & 0.32   & 0.33   \\
\textbf{ave IoU}  & 0.39    & 0.35  \\   
  \bottomrule
\end{tabular}
\end{table} 

Among five classes, delamination obtained the highest AP followed by bowl-depression and alligator crack. This occurrence is more likely due to the unique, uniform, and distinguishable appearance of these three distresses. Delaminations and bowl-shaped depressions commonly have homogeneous characteristics that are mostly circular or elongated in appearance which only varies on the type of pavement the defects are located. These two defects also have similarities in appearance leading to misclassifications by the YOLO network in which bowl-shaped depressions are labeled as delamination and vice versa. 

Cracks and scaling obtained extremely low average precision. Cracks are generally difficult to annotate as diagonal cracks tend to occupy a small proportion of the bounding box. In other words, the entirety of its bounding box is covered with almost nothing but pavement, as opposed to defects such as delamination and bowl-shaped depressions. This means that the captured object inside the anchor box provides adequate features to the network. Scaling, on the other hand, has bounding boxes that typically encase the whole image. Hence, this makes it challenging for the network to distinguish the difference between normal pavement surfaces and the defect.

\textbf{The Effect of Pre-processing}
\newline
The contrast of the image is one factor in the network's capability to detect the fault in the image. Adjusting the contrast with CLAHE increased the model's mAP and F1-score by a small margin. Four out of five distresses gained an increment on their individual average precision as these defects are better highlighted when the contrast is increased. However, the network suffered a decrease in performance on bowl-shaped depressions as this specific fault tend to have characteristics similar to delaminations when the contrast is adjusted. Figure \ref{fig:clahe} shows a comparison of the original image and pre-processed image together with predictions only generated by the network after the preprocessing is applied and vice versa. Fig \ref{fig:clahe} (a) displays the advantage of adjusting the contrast of the image to bring out distinguishable characteristics of pavement defects such as scaling. As seen in the image, there is improved visibility of disintegration on the pavement characterized by varying pixel intensities reflected by the disintegrated surface that helped in detecting that which is not observed in the original image. However, Fig \ref{fig:clahe} (b) exhibits the disadvantage of using CLAHE which removed the distinct characteristic of the fault resulting in the appearance being altered. Overall, preprocessing the image by adjusting the contrast yields better results on pavements that are more noticeable with a change in contrast but the increment in scores is negligible. YOLOv4 already employs augmentations and tricks to adapt its predicting capability on objects that are heavily influenced by brightness and contrast.

\subsection{Performance Evaluation of DeepLabv3}
\begin{figure}[h!]
  \centering
  \includegraphics[width=0.45\textwidth]{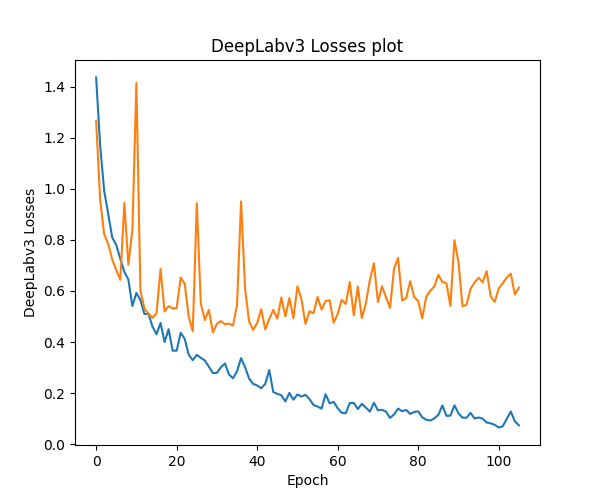}
  \caption{Plot of DeepLabv3 Loss}
  \label{fig:deeplabloss}
\end{figure}

Since YOLOv4 gained a benefit from preprocessing the image, the DeepLab model was trained on the same pre-processed images using CLAHE. As seen on Figure \ref{fig:deeplabloss}, the validation loss does not decrease anymore after around 40 to 50 epochs while the training loss keeps on decreasing. The training took 8 hours and was stopped after the 110th epoch to avoid overfitting. This phenomenon was observed in the previous experiments using YOLOv4 where the validation accuracy of the model starts to stagnate midway the training process as a clear consequence of overfitting because of the small data available. 

\subsubsection{Metrics: IoU and Dice Coefficient}
Jaccard Similarity (IoU) and dice coefficient are the metrics used to evaluate the performance of DeepLabv3 on the DPWH dataset. IoU has been discussed in the previous sections. Since this is a  multi-class segmentation, the mean IoU of the image is calculated by taking the IoU of each class and averaging them.

\textbf{Dice coefficient}. The Dice Coefficient is 2 x the Area of Overlap divided by the total number of pixels in both images. Dice coefficient is very similar and positively correlated to IoU. It is computed between the ground truth mask and the predicted mask.

\begin{centering}
$Dice = \frac{2 |X \cap Y|}{|X| + |Y|}$
\end{centering}

\subsubsection{DeepLabv3 on DPWH dataset}
Figure \ref{fig:deeplab} shows a visual output with some examples of properly classified and segmented pavement defects in form of image masks. Each distress has been colored differently for better visualization. Similar to the previous experiments using YOLO for object detection, pavement distresses such as delamination and alligator cracks are detected and segmented by the model quite well. However, unlike object detection using YOLO, DeepLabv3 was able to capture scaling and cracks on pavements even if two defects are on top of each other. The network was capable of segmenting cracks (Fig \ref{fig:deeplab} (c, d)) even with complex crack appearance. The DeepLabv3 model was able to achieve a mean IoU (mIoU) of 0.56 and a dice coefficient of 0.58 with only 200 training images. 

\begin{figure}
  \centering
  \includegraphics[width=0.5\textwidth]{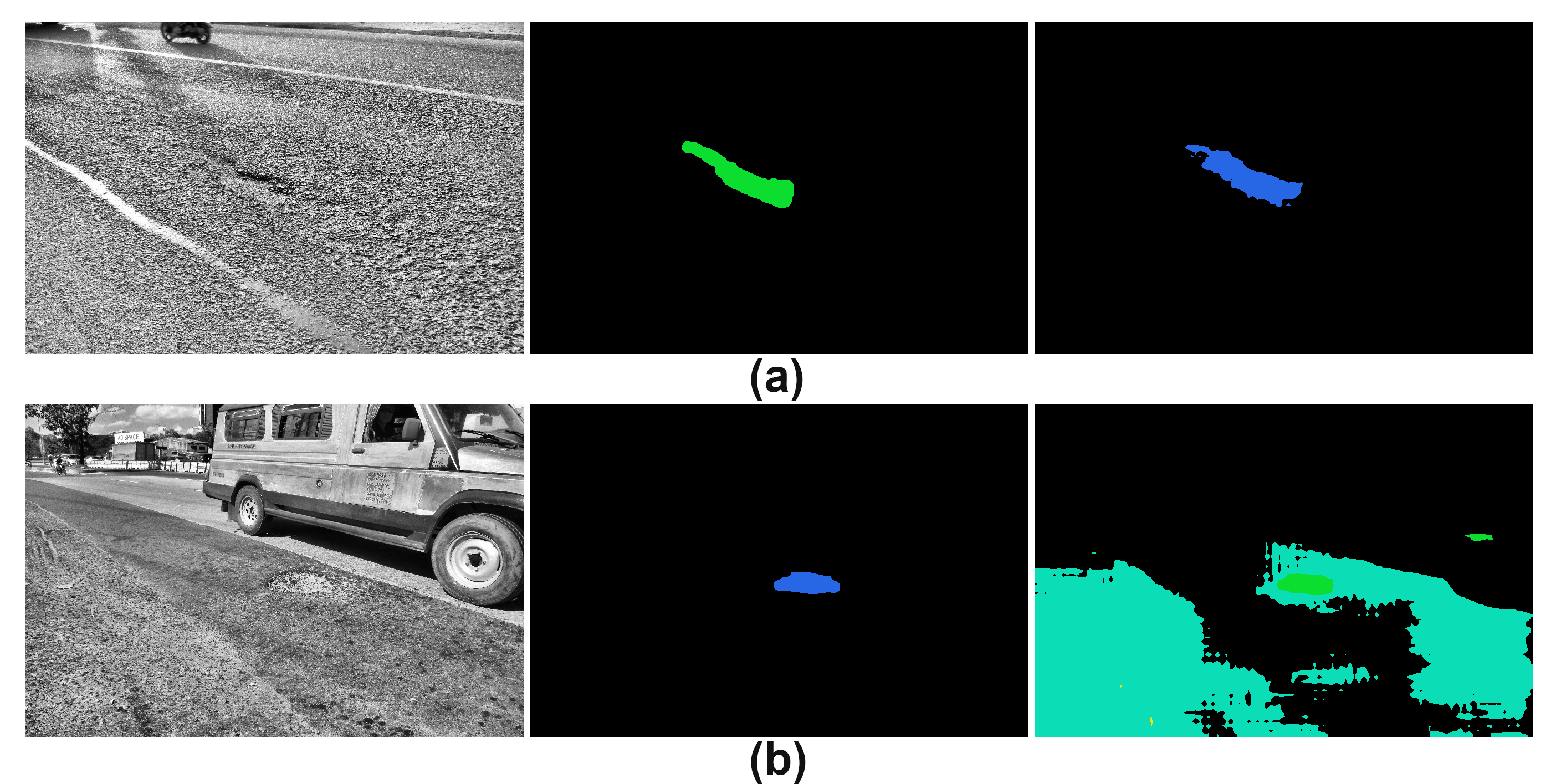}
  \caption{(a) misclassification: delamination classified as bowl-shaped depression (blue), (b) excessive segmentation: pavement classified as scaling}
  \label{fig:deeplabwrong}
\end{figure}

The semantic segmentation model misclassified delamination as bowl-shaped depression and vice versa on several occasions as observed in Figure \ref{fig:deeplabwrong}. The object detection model also suffered the same issue and might be related to the same problem when using CLAHE as it removes features related to colors in exchange for a more visible and distinct fault structure. The model is then sensitive to texture variations in the image and tends to excessively segment and classify portions of the pavements as scaling. However, the pavement on Figure \ref{fig:deeplabwrong} (b) is really close to what disintegration on the pavement surface looks like. Still, the annotations from the road engineer shall have the final verdict.

Overall, DeepLabv3 was able to do a decent job in the semantic segmentation of pavement distresses in the image despite a small training dataset and few iterations with considerable IoU and dice coeficient metrics. It was able to detect faults that are too complicated for YOLOv4 and was able to give satisfactory results in segmenting defects such as cracks and scaling. 

\section{Conclusion}
Road transport is a key infrastructure in the economic development of a country and accounts for the majority of passenger transport and cargo traffic. However, roads tend to deteriorate over time due to natural causes in the environment and repeated traffic loads. Pavement Distress (PD) detection is essential in monitoring the current conditions of the public roads to enable targeted rehabilitation and preventive maintenance. However, distress detection surveys are still manual for developing countries such as the Philippines. The use of deep learning in the detection and segmentation of pavement distresses on 2D RGB images could potentially spark solutions in building a scalable and automated end-to-end solution for PD detection in the country.

This study proposed the use of YOLOv4 in pavement distress detection and DeepLabv3 in the segmentation of pavement faults from the images of public roads in the Philippines.  While DeepLabv3 was able to achieve results of 0.56 in mIoU and 0.58 in the dice coefficient, YOLOv4 struggled on obtaining high metric scores with only 0.272 mAP on original images and 0.286 on pre-processed data. This could be mainly attributed to three reasons - having a small dataset of 305 images for five classes, non-uniform data collection, and types of distresses that are not suitable for bounding box type of detection. Both models were decent in their respective tasks of detecting and segmenting faults for delaminations, bowl-shaped depressions, and alligator cracks. PD detection with YOLOv4, however, tends to suffer from high false negatives (i.e. missed detections) on cracks and scalings while DeepLabv3 favors segmenting and classifying normal pavement surfaces as scaling yielding to relatively higher false positives on this defect. The results of this study shall provide a basis for the plausibility of creating future end-to-end solutions in automating PD detection in the country.

\subsection{Future Work}
Both object detection and semantic segmentation did not yield strong results due to the nature of the datasets. Future studies involving distress detection in pavements should consider a uniform and consistent way of collecting pavement fault datasets whether it is top-view or wide-view. Moreover, building a "one size fits all" model would have difficulty on some groups of the datasets, yielding high results on some distresses while having very low results on the others. To create an accurate generalized pavement distress detection solution, it is better to create an ensemble of models where each part specializes in a particular group or category of distresses as defects tend to vary greatly in their appearances.

\bibliographystyle{ieeetr}
\bibliography{sample}

\end{document}